\documentclass[11pt,a4paper]{article}
    \usepackage[hyperref]{emnlp2018}
    \usepackage{times}
    \usepackage{latexsym}
    \usepackage{url}
    \usepackage{CJK}
    \usepackage[utf8]{inputenc}
    \usepackage{amsmath}
    \usepackage{graphicx}
    \usepackage{subcaption}
    \usepackage{natbib}
    \usepackage{pgfplots}
    \usepackage{pgfplotstable}
    \pgfplotsset{width=6.0cm,compat=1.5}
    \usepackage{color,xcolor,framed}
    \aclfinalcopy % Uncomment this line for the final submission
    \newcommand*{\affaddr}[1]{#1} % No op here. Customize it for different styles.
	\newcommand*{\affmark}[1][*]{\textsuperscript{#1}}
    
    \title{Future-Prediction-Based Model for Neural Machine Translation}
    
    \author{
    Bingzhen Wei\thanks{Equal Contribution} \affmark[1] \quad 
    Junyang Lin\footnotemark[1] \affmark[2] \\
    \affaddr{\affmark[1]MOE Key Lab of Computational Linguistics, School of EECS, Peking 	University}\\
   	\affaddr{\affmark[2]School of Foreign Languages, Peking University}\\	
    \texttt{\{weibz, linjunyang\}@pku.edu.cn}
    }
    
    \date{}

\begin{document}
    \maketitle
    \begin{CJK}{UTF8}{gbsn}
    \begin{abstract}
    We propose a novel model for Neural Machine Translation (NMT). Different from the conventional method, our model can predict the future text length and words at each decoding time step so that the generation can be helped with the information from the future prediction. With such information, the model does not stop generation without having translated enough content. Experimental results demonstrate that our model can significantly outperform the baseline models. Besides, our analysis reflects that our model is effective in the prediction of the length and words of the untranslated content. 
    \end{abstract}

    \section{Introduction}
    \label{intro}
    
    Recent researches in machine translation focus on Neural Machine Translation, whose most common baseline is the sequence-to-sequence (Seq2Seq) model \citep{Kalchbrenner,seq2seq,ChoEA2014} with attention mechanism \citep{attention,stanfordattention,coverage, micover,interactive,multichannel,googleattention,supervisedattention,deconvdec,sact,aca}. In the Seq2Seq model, the encoder encodes the source text for a representation of the source text and decodes it for a translation that approximates the target. However, a salient drawback of this mechanism is that the decoding process should follow the sequential order, which cannot take the information in the untranslated content into consideration. Without the information about the untranslated content, the translation may end up with faults on semantic level (e.g., the translation ends by mistake with contents untranslated). The information about the ``future generation'' can provide indication for present generation, guaranteeing the loyalty of translation to the source text.
    
    To tackle the problem, we propose a novel model that targets on the provision of the untranslated information for the decoder. Based on the conventional attention-based sequence-to-sequence (Seq2Seq) model, we implement a novel decoder that is able to generate more than the present word. At each time step, the model produces a conjecture of the bag of the following words (e.g., the model is to generates a sentence “the new plan can boost the economy”, when the model generates “the new plan”, it can predict the bag of the following words that is \{can, boost, the, economy\}). Moreover, the decoder can also predict the length of the untranslated content, so as to make sure that the translation does not end without having translated all the source information. Our proposed model can be effective in generating translation with the help of the prediction of the bag of words and text length of the untranslated content. 
    
    % Moreover, to preserve the semantic meaning of the source summary, we add a semantic-relevance-based regularization term to reach this objective.

    Our contributions are summarized as below: (1). We propose a novel model for NMT that targets on the prediction of the untranslated content, which guarantees that the system can generate translation that is loyal to the source text; (2). Experimental results demonstrate that our model can significantly outperform the baseline models. (3). The analysis reflects that our model can be effective in predicting the words and text length of the untranslated content.

    \section{Model}

    % \begin{figure}[t]
    % \centering
    % \includegraphics[width=0.8\linewidth]{overview}
    % \caption{Model architecture. There are three components in the proposed model, i.e., the LSTM encoder, the deconvolutional decoder, and the conventional LSTM decoder. The encoder distills the input sentence into a vector $h_n$, which is then used in the deconvolutional decoder to obtain the global information of the target-side contexts. Based on the target-side contexts and the input-side contexts, the conventional LSTM decoder generates the output from the input-encoded vector $h_t$.}
    % \end{figure}
    
    In the following, we introduce the details of our model, including the basic attention-based Seq2Seq model and our proposed Future-Prediction-Based model.
    
    \subsection{Seq2Seq with Attention}
    In our model, the encoder, a bidirectional LSTM \citep{LSTM}, reads the embeddings of the input text sequence ${x = \{x_{1}, ..., x_{n}\}}$ and encodes a sequence of source annotations ${h = \{h_{1}, ..., h_{n}\}}$. The decoder, which is also an LSTM, decodes the final state $h_n$ to a new sequence to approximate the target with the application of conventional attention mechanism \citep{attention}. The model is trained by maximum likelihood estimation (MLE) to minimize the difference between the generation and target.

    \subsection{Future-Prediction-Based Decoder}
    
    In the following, we introduce the details our proposed future-prediction-based decoder, including the bag-of-words(BOW) predictor and the length predictor.

    \subsubsection{Bag-of-Words Predictor}
    On top of the output of the LSTM decoder, we implement a Bag-of-Words (BOW) Predictor in order to predict the word set of the following text sequence to generate. Some studies \citep{BOW_Ma} show that using Bag-of-Words as target can improve the performance of the model. With the objective of predicting the words in the future generation, the decoder can obtain more information about the target-side information. With the information about the future, it is less possible for the model to repeat the previous generation and generates translation far different from the target. Moreover, if the BOW predictor successfully predicts the word set, it can encourage the model not to generate words outside of the word set and avoids mistake. The details are in the following:
    %add
    \begin{align}
    h_{t,k} &= f_k(C_t, o_{t-1,k}) \\
    g_{t,k} &= sigmoid(h_{t,k}) \\
    z_{t,k} &= g_{t,k} \cdot tanh(C_t) + (1-g_{t,k})\cdot o_{t-1,k} \\
    o_{t,k} &= Attention(z_{t,k}, context) \\
    p_{t,k} &= softmax(Wo_{t,k}) \\
    p_t &= \frac{1}{k} {\sum^{k}_{i=1} p_{t,k}} \label{eq_pt}
    \end{align}
    where $C_t$ refers to the cell state of LSTM and ${f_k(\cdot)}$ refers to the $k$-th linear function. Since a single output is hardly able to predict all of the untranslated words, the model generates $k$ outputs for improved prediction. The averaged $p_t$ refers to the probability distribution of the untranslated words, which is used to compute the loss below.
    %add
    
    As to the representation of the target word set, we use one-hot representation by assigning $1/m$ to the word indices and 0 to others for the construction of the representation vector, where $m$ refers to the number of words in the target. Therefore, the model can be trained by minimizing negative log likelihood, where the loss $L_{BOW}$ is illustrated below:
    \begin{align}
    %L_{BOW} &= -\log P(y_{>t}|y_{<t}, x)\\
    L_{BOW} &= -\frac{1}{{N}}{\sum_{i=1}^{N}}{\sum_{t=1}^{T}}{\frac{1}{m}logP(y_{>t}^{(i)}}|{\tilde{y}_{<t}^{(i)},x^{(i)}, \theta)} 
    \end{align}
    
    With the purpose of helping the prediction at the next time step with the information about future generation, we retrieve the word embeddings of the words that the model predicts and generates a representation of the untranslated bag of words, which is shown below:
    \begin{align}
    %e_{BOW} &= \frac{1}{m}\sum_{i=1}^{m}e_i \\
    e_{bow}^t &= \sum_{i=1}^{N}p_{t-1}^i \cdot e_i
    \end{align}
    where $e_i$ refers to the word embedding, $N$ refers to the vocabulary size and $p_t$ is from \ref{eq_pt}. The $e_{bow}$ is then added to the original input for the input of the next time step.
    
    \subsubsection{Length Predictor}
    Similar to the BOW predictor, we use the output of the LSTM decoder $s_t$ as input and implement an MLP as well as softmax function on top. Also, for further information about the translation, we implement attention mechanism for the output of the current time step to extract information from the previous generation, as mentioned above. We set the length of sequence that is untranslated to a one-hot representation vector whose size is $k$, where $k$ is a hyperparameter (e.g., suppose there are still 10 words to generate according to the target text, we assign 1 to the index 10 of the vector and 0 to the others). Therefore, this still can be trained with maximum likelihood by minimizing the following loss:
    \begin{align}
    L_{len} &= -\log P(l_{y_{>t}}|l_{y_{<t}}, x)
    \end{align}
    where $l$ refers to sequence length.
    
    \subsection{Training}
    Given the parameters $\theta$ and source text $x$, the model generates a sequence $\tilde{y}$. The learning process is to minimize the negative log-likelihood, which is between the generated text $\tilde{y}$ and reference $y$, which in our context is the sequence in target language for machine translation:
    \begin{align}
    L_{NLL} &= -\frac{1}{{N}}{\sum_{i=1}^{N}}{\sum_{t=1}^{T}}{logP(y_{t}^{(i)}}|{\tilde{y}_{<t}^{(i)},x^{(i)}, \theta)} \label{eq22}
    \end{align}
    Our total loss function can be illustrated below:
    \begin{align}
    \mathcal{L} &= \lambda_1L_{NLL} + \lambda_2L_{BOW} + \lambda_3L_{len}
    \end{align}
    where $\lambda_1$, $\lambda_2$ and $\lambda_3$ are hyper-parameters. We set them to 1, 1, and 0.1 respectively in our experiments based on the model's performance on the development set. %add

    \section{Experiment}
    We evaluate our proposed model on machine translation tasks and provide the analysis.
    We present the experimental details in the following, including the introduction to the datasets as well as our experimental settings.
    
    \subsection{Datasets}
    \noindent \textbf{English-German Translation} We implement our model on the dataset WMT 2014 with 4.5M sentence pairs as training data. The news-test 2013 is our development set and the news-test 2014 is our test set. Following \citet{gnmt}, we segment the data with byte-pair encoding \citep{BPE} and we extract the most frequent 50K words for the dictionary. 
    
    \noindent\textbf{English-Vietnamese Translation} Following \citet{luong2015stanford}, we use the same preprocessed data for this task with 133K training sentence pairs \citep{2015iwslt} for training. The TED tst2012 with 1553 sentences and the the TED tst2013 with 1268 sentences are our development and test set respectively. We preserve casing, and we set the English dictionary size to 17K words and Vietnamese dictionary to 7K words. The case-sensitive BLEU score \citep{bleu} is the evaluation metric.

    \begin{table}[tb]
    \centering
        \begin{tabular}{l | c}
        \hline
        Model & BLEU  \\ \hline\hline
        ByteNet  & 23.10                 \\
        GNMT & 24.60\\
        ConvS2S  & 25.16 \\ \hline\hline
        Seq2Seq (our reimplementation) & 25.14 \\
        \textbf{FPB} &  \textbf{25.79} \\
        \hline
        \end{tabular}
        \caption{\textbf{Results of the models on the English-German translation.} }
        \label{ende}
    \end{table}
    
    \subsection{Setting}
    We implement the models in PyTorch on an NVIDIA 1080Ti GPU. Both the size of word embedding and the number of units of hidden layers are 512, and the batch size is 64. We use Adam optimizer \citep{KingmaBa2014} with the default setting, $\alpha=0.001$, $\beta_1=0.9$, $\beta_2=0.999$ and $\epsilon=1\times10^{-8}$, to train the model. Gradient clipping is applied with the norm smaller than 10. Dropout \citep{dropout} is used with the dropout rate set to 0.2 for both datasets, in accordance with the model's performance on the development set. Based on the performance on the development set, we use beam search with a beam width of 10 to generate text.
    
    \subsection{Baselines}
    For the English-German translation, we compare with the baseline models in the following. \textbf{ByteNet} is the Seq2Seq model based on dilated convolution, which runs faster than conventional RNN-based model \citep{bytenet}. \textbf{GNMT} is the improved version of end-to-end translation system that tackles many detail problems in NMT \citep{gnmt}. \textbf{ConvS2S} is the Seq2Seq model completely based on CNN and attention mechanism, which achieves outstanding performance in NMT. 
    
    For English-Vietnamese translation, the models to compared are presented below. \textbf{RNNSearch} The attention-based Seq2Seq model as mentioned above, and we present the results of \citep{luong2015stanford}.
    
    For both datasets, we reimplement the baseline, the attention-based Seq2Seq model, which is named \textbf{Seq2Seq}.
    
    \section{Results and Analysis}
    In the following, we present our experimental results as well as our analysis of our proposed modules to figure out how it enhances the performance of the basic Seq2Seq model for NMT.

    \begin{table}[tb]
    \setlength{\tabcolsep}{2.5pt}
    \centering
        \begin{tabular}{l | c}
        \hline
        Model & BLEU  \\ \hline\hline
        RNNSearch &     26.10                 \\ \hline\hline
        Seq2Seq (our reimplementation) & 25.90\\
        \textbf{FPB} &  \textbf{27.70} \\
        \hline
        \end{tabular}
        \caption{\textbf{Results of the models on the English-Vietnamese translation.} }
        \label{envi}
    \end{table}
    \subsection{Results}
    Table \ref{ende} shows the results of our model as well as the baseline models on the English-German translation dataset.
    
    Table \ref{envi} shows the results of the models on the English-Vietnamese translation dataset. It can be found that on the evaluation of BLEU score, our proposed model has significant advantage over the RNNSearch, which demonstrates that our proposed model is effective in improving the performance of the baseline. In the following, we conduct ablation test to evaluate the effect of each module and examine the performance of the BOW predictor in prediction accuracy of words.
    
    \subsection{Ablation Test}
    To evaluate the effects of each proposed module, we conduct an ablation test for our model to examine the individual effect of our BOW predictor and length predictor.
    
    We present the results of the ablation test on Table~\ref{ablation}. Compared with the basic attention-based Seq2Seq model, it can be found that the length predictor can bring a slight improvement for the baseline model, while the model only with the BOW predictor can outperform the baseline with a large margin. It is obvious that the BOW predictor brings contribution to the model's performance, and we analyze its bag-of-words prediction accuracy in the next section. The combination of the two modules, which is our proposed model, can achieve the best performance.
    
    \begin{table}[tb]
    \setlength{\tabcolsep}{2.5pt}
    \centering
        \begin{tabular}{l| c}
        \hline
        Model & BLEU  \\ \hline\hline
        Seq2Seq (our reimplementation)  &     25.90                 \\ 
        +length predictor & 26.26 \\
        +BOW predictor & 27.38 \\
        \textbf{FPB} &  \textbf{27.70} \\
        \hline
        \end{tabular}
        \caption{\textbf{Ablation test on the English-Vietnamese translation.} Seq2Seq refers to our reimplementation of the attention-based Seq2Seq model}
        \label{ablation}
    \end{table}

    \subsection{Bag-of-Words Prediction}

    In this section, we present our analysis of the prediction accuracy of the BOW predictor. As the BOW predictor predicts words at each decoding time step, we evaluate its accuracy in various situations by evaluating its bag-of-words prediction accuracy with different lengths of untranslated words. For example, if there are still 20 words left for translation, we evaluate if the BOW predictor can predict the correct words without concerning sequential order. 
    
\begin{figure}[tb]
    \centering
    \includegraphics[width=1.0\linewidth]{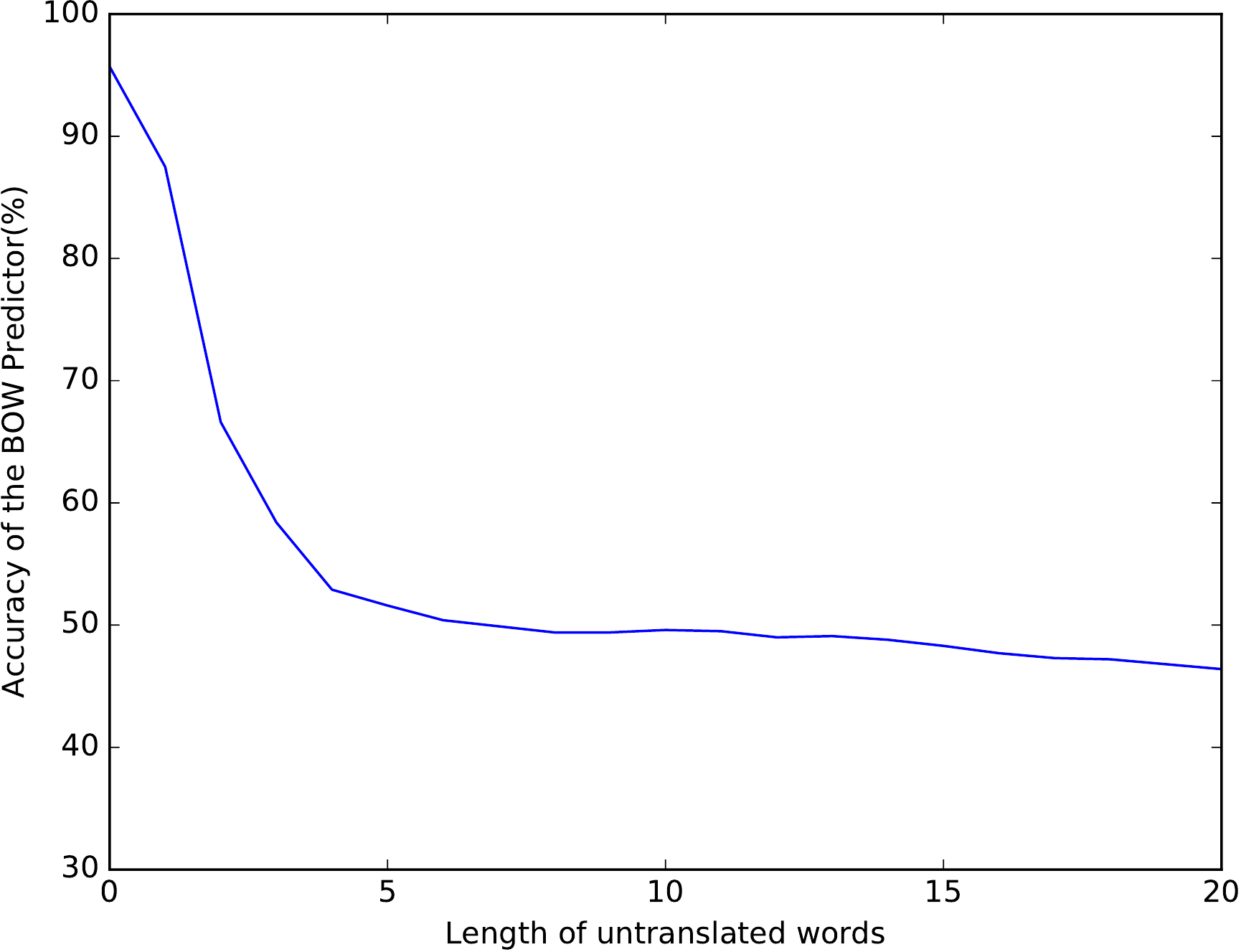}
    \caption{Accuracy of the BOW prediction at the time step with different lengths of untranslated words}
    \label{acc}
    \end{figure}

    Results shown in Figure~\ref{acc} reflect our model's performance on the prediction of the bag of words to translate at different time steps with diverse lengths of untranslated content. It can be found that with the increase of untranslated words, the prediction accuracy decreases. The phenomenon is reasonable as it is more difficult to predict the information about further future only with the information from the source-side context and the previous generation. However, even when the length of the untranslated words is relatively long (20 words), the model can still maintain a stable performance on the evaluation with the accuracy of around 50\%. This demonstrates that our model possesses strong capability of predicting the word-level information about future generation.
    
    \section{Conclusion and Future Work}
    \label{conclusion}
    In this paper, we propose a novel model for NMT with the BOW predictor that predicts the words that are not translated and the length predictor that predicts the length of the untranslated words. Therefore, the model can receive information about the future from its conjecture to improve the quality of the current translation. Experimental results demonstrate that our model outcompetes the baseline model on the English-Vietnamese translation dataset. Moreover, our analysis shows that our proposed modules can enhance the performance of the baseline individually, especially the BOW predictor, and we find that the BOW is able to predict words with high accuracy and the accuracy increases with the decline of the number of untranslated words.

    \bibliography{emnlp2018}
    \bibliographystyle{acl_natbib_nourl}
    
    \end{CJK}
    
    \end{document}